**Empirical Study of Large Language Models as Automated Essay Scoring Tools in English Composition——Taking TOEFL Independent Writing Task for Example**

Wei Xia[1] , Shaoguang Mao[2] , Chanjing Zheng[1,3]

Author Note

[1] Department of Educational Psychology of Faculty of Education, East China Normal University, Shanghai, China;

[2] Microsoft Research Asia, Beijing, China;

[3] Shanghai Institute of Artificial Intelligence Education, East China Normal University, Shanghai, China.

Correspondence concerning this article should be addressed to Chanjing Zheng, Department of Educational Psychology of Faculty of Education and Shanghai Institute of Artificial Intelligence Education, East China Normal University, Shanghai, 200062; China. Email: chjzheng@dep.ecnu.edu.cn

**Abstract**

Large language models have demonstrated exceptional capabilities in tasks involving natural language generation, reasoning, and comprehension. This study aims to construct prompts and comments grounded in the diverse scoring criteria delineated within the official TOEFL guide. The primary objective is to assess the capabilities and constraints of ChatGPT, a prominent representative of large language models, within the context of automated essay scoring.

The prevailing methodologies for automated essay scoring involve the utilization of deep neural networks, statistical machine learning techniques, and fine-tuning pre-trained models. However, these techniques face challenges when applied to different contexts or subjects, primarily due to their substantial data requirements and limited adaptability to small sample sizes. In contrast, this study employs ChatGPT to conduct an automated evaluation of English essays, even with a small sample size, employing an experimental approach.

The empirical findings indicate that ChatGPT can provide operational functionality for automated essay scoring, although the results exhibit a regression effect. It is imperative to underscore that the effective design and implementation of ChatGPT prompts necessitate a profound domain expertise and technical proficiency, as these prompts are subject to specific threshold criteria.

*Keywords:* ChatGPT, Automated Essay Scoring, Prompt Learning, TOEFL Independent Writing Task

**Empirical Study of Large Language Models as Automated Essay Scoring Tools in English Composition——Taking TOEFL Independent Writing Task for Example**

## Introduction

Writing practice plays a pivotal role in the language acquisition process. With a growing number of students, the workload for educators tasked with reviewing student essays has surged significantly. Consequently, Automated Essay Scoring (AES) methods have gained widespread popularity all around the world. AES systems offer the potential to expedite the essay correction process, furnish feedback notes, and assist students in identifying crucial elements of writing, thereby enhancing their writing proficiency. At the heart of automated essay scoring systems lies the core technology known as Natural Language Processing (NLP), an interdisciplinary field encompassing linguistics, computer science, and related domains. NLP investigations encompass diverse linguistic units, such as characters, words, phrases and paragraphs, along with a spectrum of activities, including natural language processing, comprehension, and generation. Given the prevailing state of natural language technology at the time of this research, automated essay scoring systems are categorized into five developmental stages.

Historically, the development of automated essay scoring technology has been largely contingent on advancements in computer technology, predating the introduction of ChatGPT (Chat Generative Pretrained Transformer). Although significant technical progress has led to an incremental improvement in the quality of automated essay scoring, several significant challenges persist. Existing research methods demonstrate subpar performance in scenarios with limited sample sizes, and the transferability of these systems remains deficient. Moreover, each task necessitates the maintenance of a distinct model, resulting in diminished model utilization efficiency. Conversely, the proliferation of model parameters escalates the costs associated with feature extraction and fine-tuning.

The debut of ChatGPT in November 2022 generated considerable intrigue. ChatGPT, developed by OpenAI, leverages supervised learning and reinforced learning with human feedback (RLHF), building on the foundation laid by earlier iterations of the GPT series, specifically GPT-3.5 or 4.0. ChatGPT possesses the capability to comprehend human language by discerning pertinent information and engaging in conversations with users to glean a deeper understanding of their intent. Thanks to its technological advancements, ChatGPT has exerted a profound influence on the realm of natural language processing, initiating a new era of automated essay scoring systems based on generative pre-trained models.

In this study, the authors employ ChatGPT to conduct automated scoring of provided essays, while concurrently formulating prompts and feedback based on the discrete writing elements outlined in the seventh edition of the official TOEFL guide, "The Official Guide to the TOEFL iBT Test" (Educational Testing Service, et al., 2021). By leveraging the outcomes of scoring and the creation of specific assessment metrics during the experimental phase, this research elucidates both the capabilities and constraints of ChatGPT in the domain of automated essay scoring.

## Overview of the Development of Automated Essay Scoring Systems from the Perspective of Natural Language Processing

The development of Automated Essay Scoring (AES) systems spans over half a century, originally conceived with a focus on psychology and academic assessment. At its core, the key technical underpinning of AES is Natural Language Processing (NLP). The emergence of AES coincided with pivotal advancements in NLP technology.

      Lim et al. (2021) categorized AES systems into three principal frameworks: the Content Similarity Framework (CSF), the Machine Learning Framework (MLF), and the Hybrid Framework, which integrates the strengths of both approaches. In line with "Handbook of Automated Scoring" (Duanli et al., 2021), NLP's involvement in AES can be delineated into three stages: early rule-based methodologies, mid-stage statistical techniques, and late-stage deep learning techniques. Liu et al. (2021) identified three major stages in the evolution of NLP: the initial fully supervised learning phase, heavily reliant on feature engineering; the subsequent transition to feature extraction through deep neural networks; and, commencing in 2021, the shift from "pre-training + fine-tuning" to "pre-training + prompt + prediction." "Introduction to Natural Language Processing" (Zhang et al., 2022) elaborates on four phases: rule-based methods, machine learning-based techniques, deep learning-based techniques, and large-scale model-based approaches.

      This section elucidates the evolution of AES from the NLP perspective. This viewpoint explores the nexus between AES and NLP, along with the merits and demerits of diverse NLP systems. The progression of NLP is partitioned into five phases, as posited by this investigation: rule-based, feature-based statistical machine learning, deep neural network-based, pre-training + fine-tuning-based, and generative large language pre-training model-based.

      The prodigious language model ChatGPT stands poised to advance the fifth phase, grounded in large-scale pre-trained language models. Consequently, our research emphatically advocates the bifurcation of AES into five distinct stages. Recognizing that the fifth stage remains in its nascent stages and draws its foundation from large-scale pre-trained language models, this endeavor may be perceived as an endeavor to incorporate this technology into the realm of AES. The succeeding sections elaborate upon these five stages.

**(1) From 1950 to 1990: The Stage of Rule-Based Research**

The rule-based approach employs linguistic vocabulary and grammatical rules to impart linguistic knowledge and execute tasks requiring natural language processing. It encompasses data creation, rule development, their implementation, and the assessment of their effects. The core idea is to offer rule formats that enable linguists to translate knowledge into rules without the need for proficiency in computer programming.

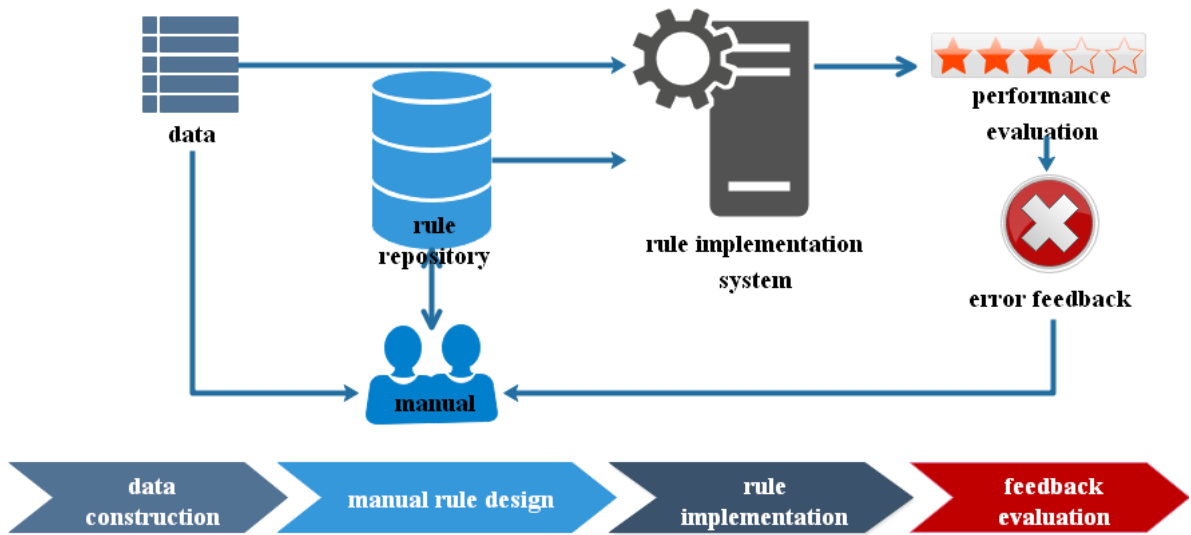

**Fig. 1 The central process of rule-based natural language processing**

Project Essay Grade (Page, 1966) stands as the pioneering automated essay scoring system at this stage. By establishing feature rules, this method primarily retrieves pertinent rules from a repository of training articles. Subsequently, it translates these rules and applies them to multiple regression, utilizing the levels present in the training articles to determine regression coefficients and provide scores.

**(2) From 1990 to 2010: The Stage of Feature-based Statistical Machine Learning**

The technological underpinning of this stage transitioned towards feature-based statistical machine learning as computer technology advanced and interest in Automated Essay Scoring (AES) grew. During this period, the predominant approach in natural language processing involved the creation of feature representations based on task-specific characteristics. Models were trained using extensive annotated corpora, and natural language processing tasks were reformulated into classification problems employing supervised classification methods. This technique encompasses artificial feature engineering, core natural language processing algorithms, appropriate machine learning models, learning criteria, and optimization algorithms for raw data analysis and information extraction.

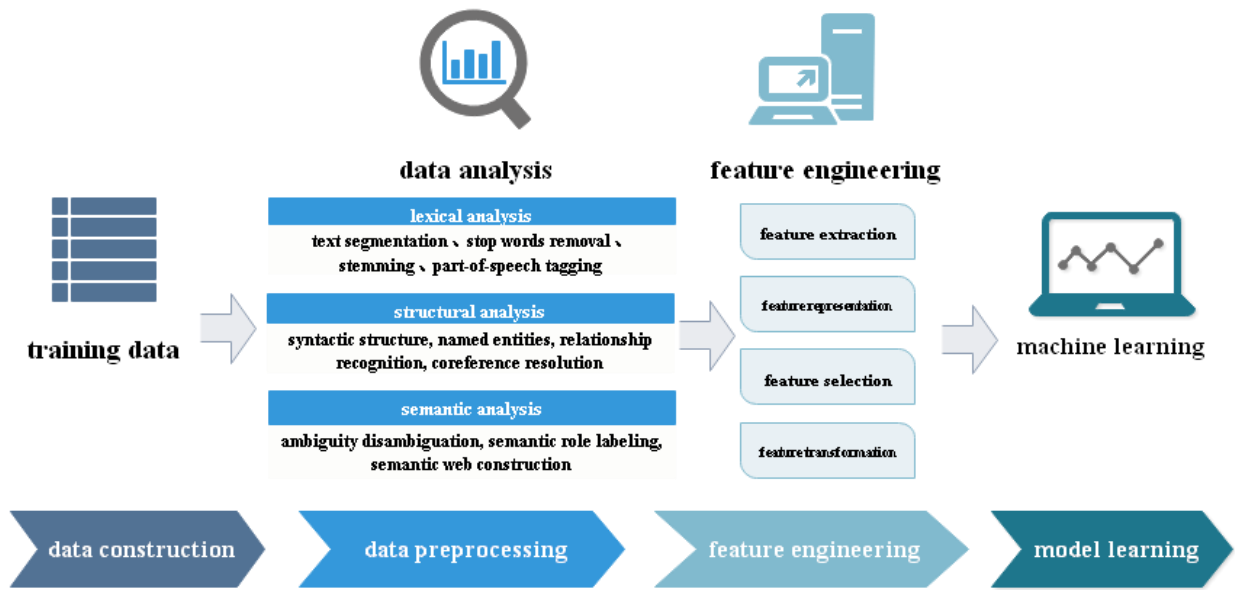

**Fig. 2 The fundamental methodology of feature-based statistical machine learning**

Consequently, in the late 1990s and early 2000s, AES systems such as the Intelligent Essay Assessor (Landauer et al., 1998) and E-raters (Attali & Burstein, 2006) were introduced. This stage predominantly relies on feature-based statistical machine learning techniques, real essay data, and corpus-based modeling approaches. It involves preprocessing article data, selecting and balancing predictive features for essay scoring based on analytical findings, and computing final scores using machine learning models.

**(3) From 2010 to 2018: The Stage of Deep Neural Network**

In the backdrop of limitations imposed by the evaluation features employed in the feature-based statistical machine learning phase, which are constrained to surface features extraction and often fail to capture the multi-dimensional and profound attributes considered by human assessors during actual essay evaluation, the research paradigm rooted in deep neural networks predominantly seeks to construct models endowed with specific "depth." This approach amalgamates feature acquisition and predictive modeling, culminating in algorithmic enhancements that facilitate the automatic acquisition of improved feature representations by the models. This stage encompasses activities encompassing data generation, preprocessing, and model training. The fundamental principle underlying this stage lies in the transformation of raw data into more abstract representations through multi-layered feature transformations, thereby permitting the acquired representations to, to some extent, supplant intentionally crafted features.

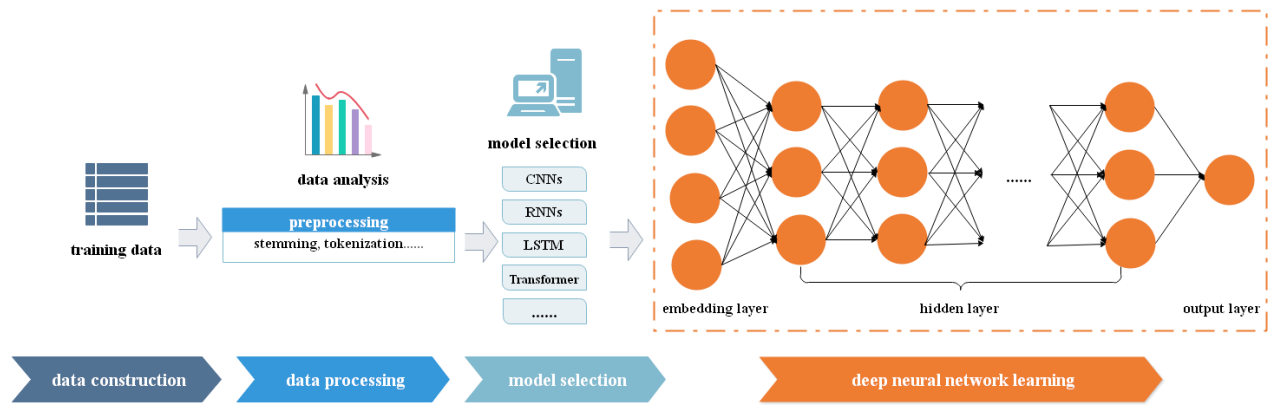

**Fig. 3 Illustrates the core mechanics of employing deep neural networks**

A pioneering instance of leveraging Convolutional Neural Networks (CNN) and Recurrent Neural Networks (RNN) in the construction of Automated Essay Scoring (AES) systems can be attributed to Taghipour and Ng (2016). Subsequently, Dong and Zhang (2016) harnessed word-level CNN+Pooling for holistic text encoding, founded upon sentence-level encoding, emulating the hierarchical structure inherent to essays. They further advanced their approach by introducing the Attention mechanism, emphasizing the superiority of CNN in sentence-level modeling, while highlighting the aptitude of RNN in comprehensive text modeling.

**(4) From 2018 to 2022: The Stage of Pre-training + Fine-tuning**

The year 2018 marked a significant milestone in the evolution of natural language processing with the introduction of the ELMo (Embeddings from Language Models) model by Peters et al. (2018). This innovation introduced the "pre-training + fine-tuning" paradigm, which revolutionized deep neural network-based natural language processing. Through pre-training models on self-supervised tasks and acquiring more generalized language representations from vast corpora, this stage fine-tunes pre-trained models to address challenges such as slow training, low efficiency, overfitting, ultimately achieving impressive performance.

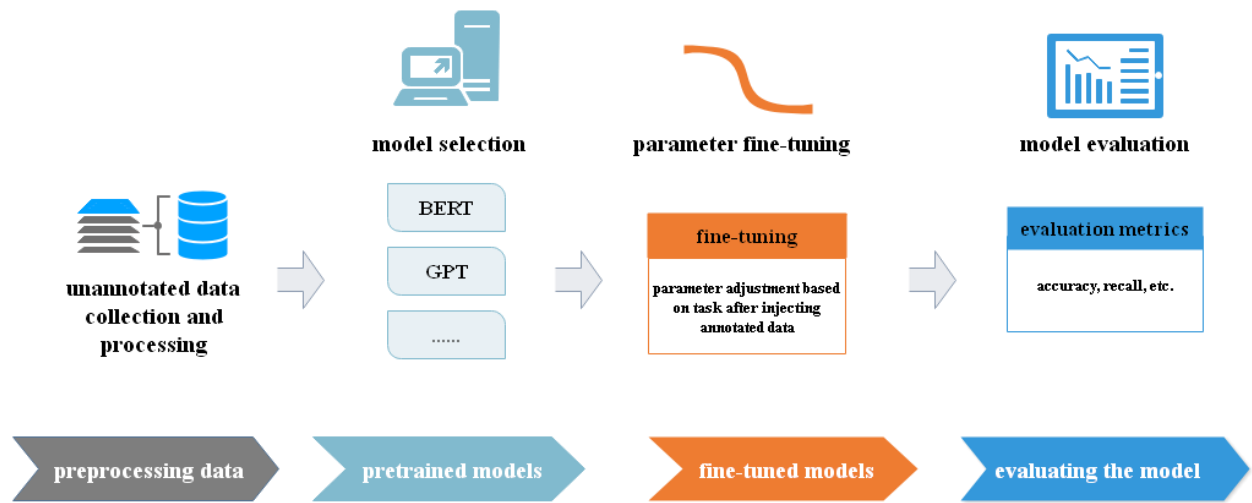

**Fig. 4 The Basic Pretraining + Fine-tuning Process for Natural Language Processing**

Devlin et al. (2019) further advanced this concept by proposing a method that simultaneously conditions all network layers. They employed this approach to pre-train deep bidirectional feature representations of unlabeled text, exemplified by the expansive language model, BERT (Bidirectional Encoder Representations from Transformers). Consequently, optimizing the BERT model became notably streamlined, requiring only a single output layer.

The intelligent Chinese composition tutoring system, developed by Zheng et al. (2023) utilizes the large language model RoBERTa an enhanced version of the BERT, to achieve automatic essays scoring. This breakthrough results in models ideally suited for various tasks, including question-answering and linguistic reasoning, without the need for substantial task-specific design modifications.

**(5) 2022 to Present: The Stage of Generative Large Language Models**

Despite the significant success achieved by the "pre-training + fine-tuning" approach in natural language processing, it still confronts challenges such as the need to develop multiple models and the high cost associated with fine-tuning. The landscape of natural language processing witnessed a substantial transformation following the release of ChatGPT in November 2022. This innovation, facilitating interactive engagement with users through prompts, opened the door to a wide array of natural language processing tasks, contingent upon specific directives.

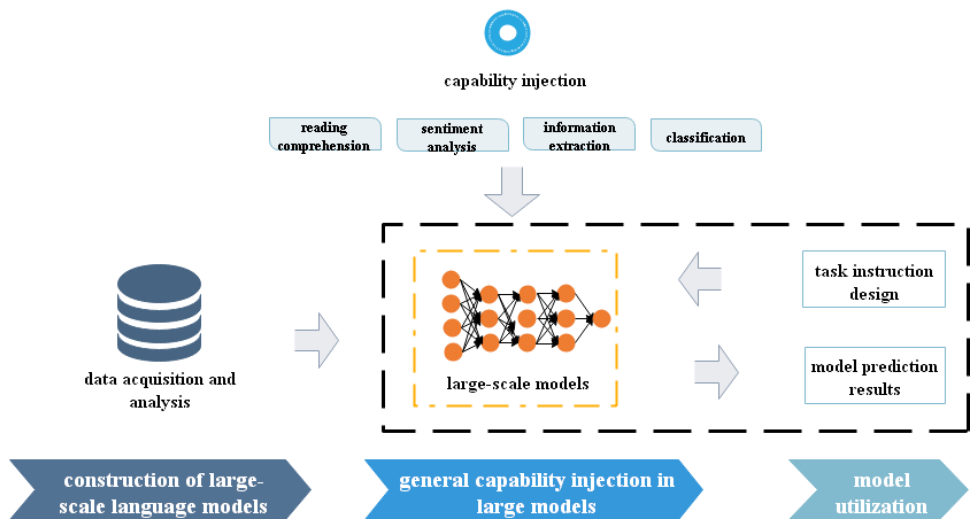

**Fig. 5 The Fundamental Process of Using Generative Large Language Models**

This paper employs the large-scale pre-trained model, ChatGPT, as a case study. The methodology involves generating contextually relevant prompt instructions based on task requirements, integrating them with the text to be evaluated, and employing user interaction to directly address downstream tasks using the pre-trained model. This approach empowers zero-shot or few-shot learning, minimizing the reliance on extensive downstream task data. Consequently, the design and implementation of task prompt instructions emerge as pivotal factors, given their profound impact on the precision of ChatGPT's essay evaluation.

## Introduction to Prompt Learning

The cornerstone of ChatGPT's interactive engagements with users lies in the concept of prompt learning. This entails the development of appropriate prompt task instructions, tailored to the specific task requirements, ensuring alignment between the provided instructions and the textual content necessary for task execution. This process can be conceptualized as a form of rapid learning.

### (1) General Prompt Learning Processes

The fundamental prompt learning process comprises four key stages: prompt creation, answer construction, response prediction, and answer-label mapping. To elucidate each of these stages, we will employ the prominent natural language processing task of sentiment classification as an illustrative example (Liu et al., 2021).

Description of the Sentiment Classification Task:

Given a sentence denoted as 'x' (e.g. "x" representing "I like this little dog"), the task involves predicting the emotional polarity of 'x' distinguishing between categories such as 'cute' and naughty.'

**1) Transformation of Input 'x' into Prompted 'X' via Prompt Construction.** Input slots, denoted as '[x]' and response slots, denoted as '[z]' represent two tasks for which a versatile text format has been devised. The data originating from 'x' is inserted into the input slot '[x]' to predict the response slot 'z' The specific procedure is depicted in the following diagram:

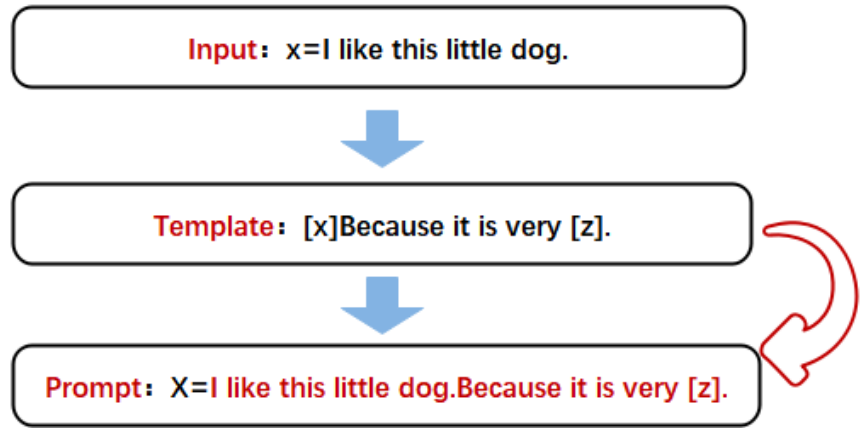

Fig. 6 Illustration of the prompt construction process (Liu et al., 2021)

**2) Answer Design: Establishing a Function for Mapping Answers to Labels.** In certain circumstances, data annotations may exhibit variations. One such scenario involves the synchronization of answers and labels, as exemplified in machine translation initiatives. Alternatively, multiple responses may correspond to a single label. For instance, in sentiment classification tasks, words such as "excellent" , "good" and "wonderful" can all be associated with a positive emotion label.

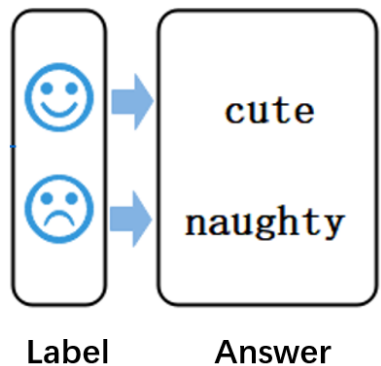

Fig. 7 Relationship Mapping Between Answers and Labels (Liu et al., 2021)

3） **Predicting the Response to a Given Question Using the Appropriate Language Model.** An appropriate language model is initially selected. By constructing input sequences denoted as X with prompts, the well-trained language model is then employed to promptly predict the output result for the answer slot [z].

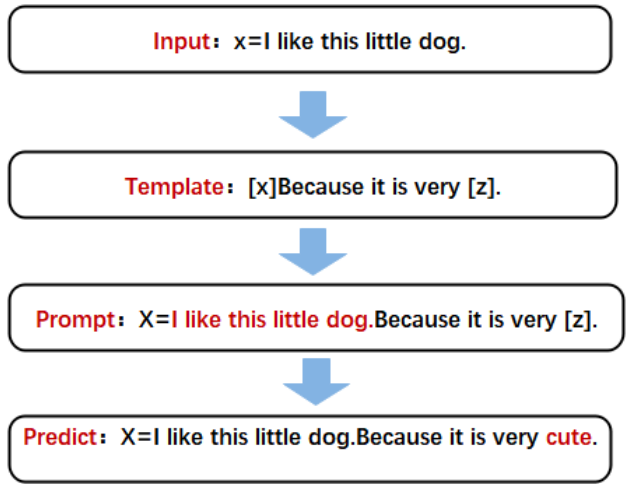

Fig. 8 Process of Answer Prediction (Liu et al., 2021)

4） **Answer-Label Mapping: Mapping the Predicted Answer [z] to Labels.**

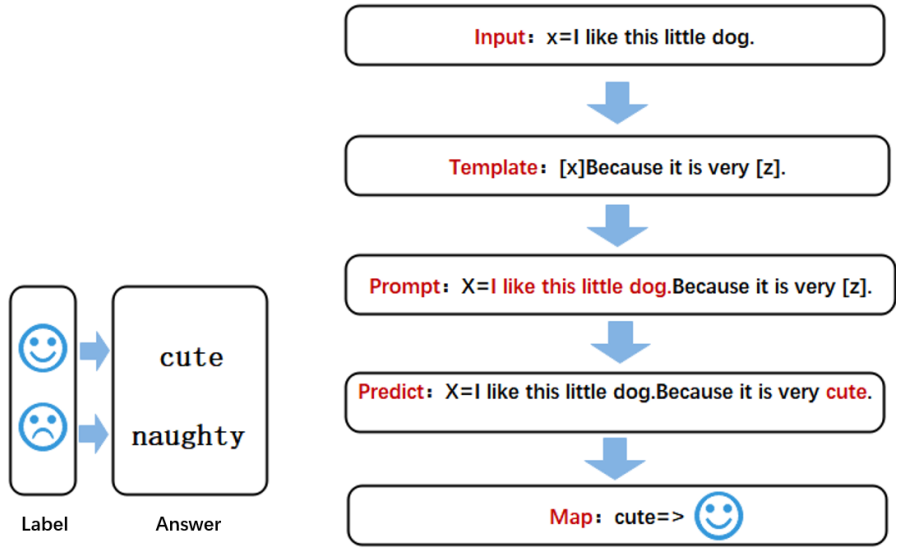

Fig. 9 Illustrates the comprehensive process involved in Sentiment Classification Tasks using the Prompt Learning Method (Liu et al., 2021)

At this juncture, the application of prompt learning has successfully accomplished a straightforward sentiment classification task.

**(2) Advantages of Prompt Learning**

ChatGPT stands as a formidable pre-trained model, harnessing the synergy of prompt learning and feedback learning to elevate its conversational prowess. This synthesis imbues it with greater user-friendliness, extending its applicability across a wide spectrum of tasks. Prompt learning equips ChatGPT with the capability to generate coherent responses within specific domains, establishing it as a pivotal player in the landscape of natural language processing.

The advantages of prompt learning in natural language processing. Prompt learning serves as a pivotal transformation in the field of Natural Language Processing (NLP), shifting it towards the realm of large-scale pre-trained language model tasks. The initial four stages of NLP tasks necessitate the development of specific task-oriented data and models. Even the fine-tuning process mandates retraining, with inherently limited exploration of few-shot learning and generalization capabilities when confronted with novel tasks. The introduction of expansive language models, such as ChatGPT, serves as a prime example, showcasing remarkable performance in text generation and adaptability to unfamiliar tasks. With the integration of prompt learning, these models exhibit a heightened ability to excel in few-shot scenarios. This is achieved by facilitating the application and transfer of knowledge within NLP tasks, consequently enhancing the overall performance of few-shot learning.

Compared to pre-training + fine-tuning, prompt learning serves as a pivotal transformation in the field of Natural Language Processing (NLP), shifting it towards the realm of large-scale pre-trained language model tasks. The initial four stages of NLP tasks necessitate the development of specific task-oriented data and models. Even the fine-tuning process mandates retraining, with inherently limited exploration of few-shot learning and generalization capabilities when confronted with novel tasks. The introduction of expansive language models, such as ChatGPT, serves as a prime example, showcasing remarkable performance in text generation and adaptability to unfamiliar tasks. With the integration of prompt learning, these models exhibit a heightened ability to excel in few-shot scenarios. This is achieved by facilitating the application and transfer of knowledge within NLP tasks, consequently enhancing the overall performance of few-shot learning.

## Experimental Design

The primary objective of this empirical study is to assess the significance of various prompt elements in the automated scoring of TOEFL independent writing tasks. Additionally, the study aims to explore the capabilities and limitations of employing ChatGPT in the domain of automated essay scoring.

The empirical study is divided into two main segments. The first segment involves prompt design. For this purpose, we draw from the TOEFL independent writing tasks section found in "The Official Guide to the TOEFL iBT Test" (Educational Testing Service, 2021). We configure the prompt information to enable ChatGPT to evaluate TOEFL independent writing tasks. We subsequently assess ChatGPT's performance under various prompt information configurations by analyzing the scores assigned by ChatGPT and computing evaluation metrics. The second segment encompasses feedback experiments based on the initial prompt design. Utilizing feedback obtained from evaluations, we conduct secondary scoring and make predictions concerning high-scoring articles on different topics within the TOEFL independent writing dataset.

The subsequent sections will provide a comprehensive explanation of the experimental content, including data acquisition, prompt design, feedback design, and evaluation criteria.

**(1) Acquisition of Experimental Objects**

To conduct the preliminary experiment, we employed TOEFL independent writing tasks extracted from "The Official Guide to the TOEFL iBT Test 6th Edition." We utilized various combinations of prompt information to automate the scoring process using ChatGPT. Following the preliminary experiment, we sought to validate the accuracy of our initial findings. For this purpose, we selected 32 high-scoring TOEFL independent writing articles, encompassing a range of 11 distinct topics. Each topic consisted of a randomly chosen assortment of articles.

Leveraging the prompt information that exhibited optimal predictive performance in the preliminary experiment, we employed ChatGPT to score articles across these diverse topics. This approach enabled us to conduct a more comprehensive assessment of the accuracy of our earlier conclusions. The ensuing table provides the specific count of predicted articles for each individual topic:

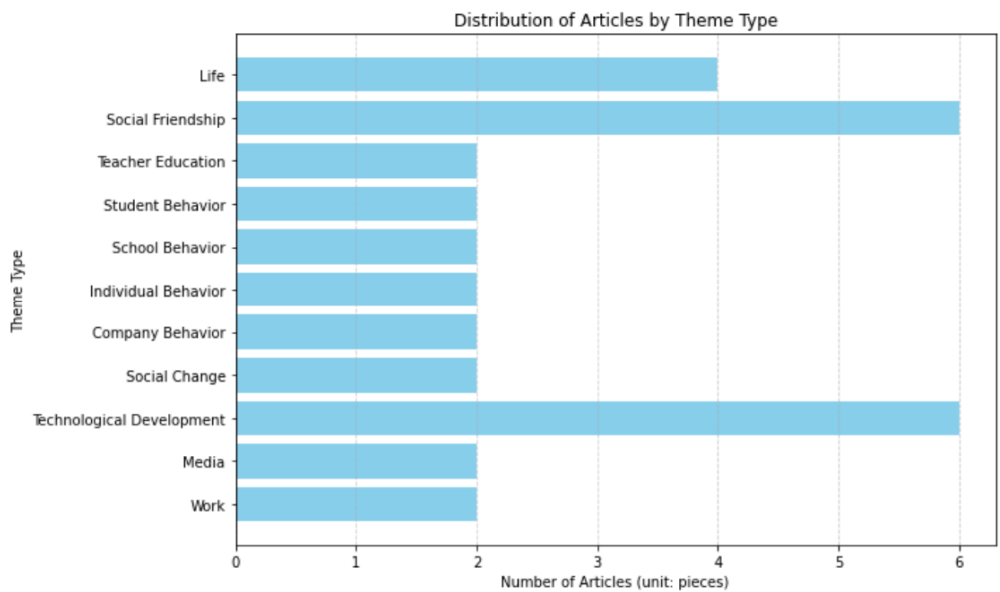

**Fig. 10 Distribution of the Number of Test Articles for Different Themes**

**(2) Prompt Design**

In the preliminary experiment, prompt information is categorized into four primary sections, aligning with the official guide's content:

- **Basic Task Description:** This section includes information pertaining to the test type and the corresponding Writing task as outlined in the official guide.
- **Scoring Criteria:** This corresponds to the Independent Writing Scoring Rubric provided in the official guide.
- **Sample Essays:** This section corresponds to the "Score X Essay" in the official guide, where 'X' represents the standard score.
- **Feedback Comments on Examples:** This mirrors the content found in the official guide under "Rater Comments."

The overarching principle guiding prompt design is to initially construct general prompt information which might be vague in instructions and subsequently refine its elements. This approach is based on the predictive outcomes generated by ChatGPT. We examine how automated scoring by ChatGPT evolves under different prompt designs and analyze the significance of various elements in the context of applying ChatGPT to automated English essay scoring.

In the composition of prompt information combinations, the fundamental task description is deemed indispensable. The components subject to combination encompass scoring rules, sample essays, and feedback comments. Given the distinctiveness of sample essays, automated essay scoring by ChatGPT may be influenced by variables such as the quantity of examples and their corresponding score levels.

To account for both the scores assigned to examples and their score levels, we partition the number of provided examples. This division initially categorizes examples into three groups: single example, double examples, and triple examples. In the case of single and double examples, a more detailed stratification is essential, considering high, medium, and low-level examples. These align with 5-point examples, 3-point examples, and 1-point examples, respectively. For single and double examples, each category is further segmented into three distinct scenarios.

In contrast, when working with three examples, the prompt information only necessitates high, medium, and low-quality examples. The ensuing section outlines the specific prompt information designed for the experiment:

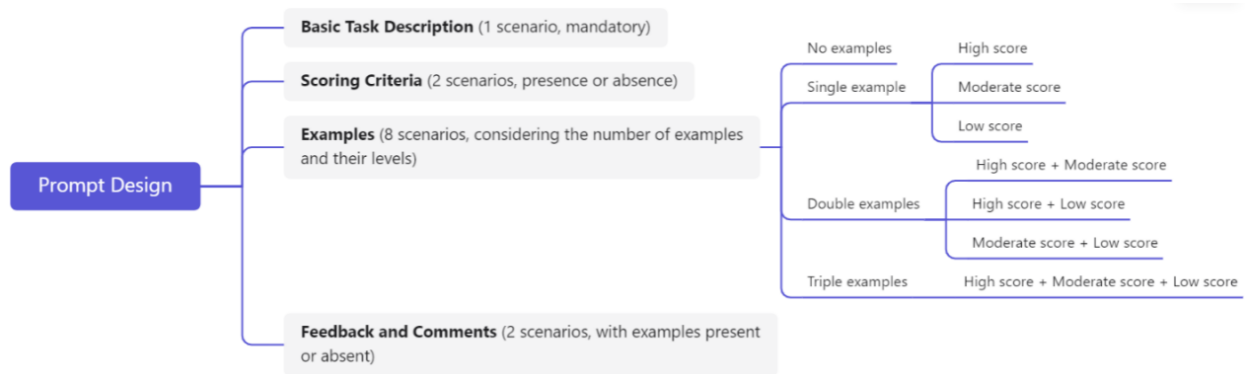

**Fig. 11 Specific Composition of Prompt Design (30 Scenarios)**

The 30 scenarios enumerated above delineate the compositional aspects of prompt information design employed in this experiment.

Taking 'Basic Task Description + Predicted Score for Single Example (3-Point)' as an illustrative instance:

1) **Prompt Construction.** This phase entails explicating the formulation of diverse prompt information during the design process. The chosen configuration for prompt information composition is 'Basic Task Description + Single Example with a 3-Point Score,' as demonstrated below:

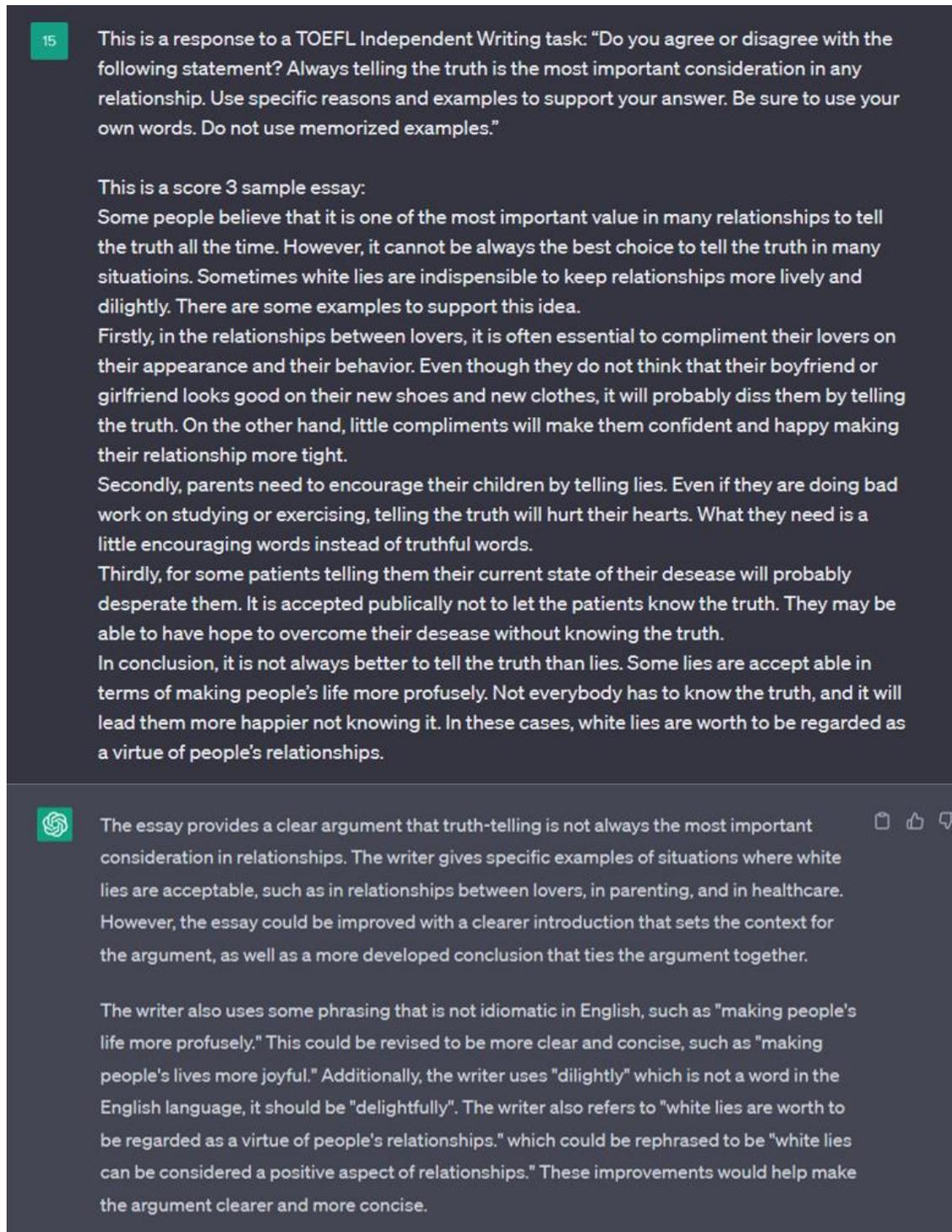

**Fig. 12 Demonstration example.**

2) **Answer Construction.** Designing the mapping function between answers and labels.

In this phase, the primary objective pertains to the anticipation of essay scores, with the mapping relationship between answers and labels being exclusively delineated within the official guidelines. Specifically, this relationship is established by the correspondence between the standardized scores allocated to predicted essays and the scores forecasted by ChatGPT, predicated upon prompt information. This correspondence is fundamentally confined to numerical values.

3) **Answer Prediction.** Utilizing ChatGPT to forecast scores based on the provided prompt information.

As illustrated below: ChatGPT assesses an essay with a standard score of 4 in accordance with the prompt information extracted from the TOEFL official guide. Furthermore, the score predicted by ChatGPT also precisely with a score of 4.

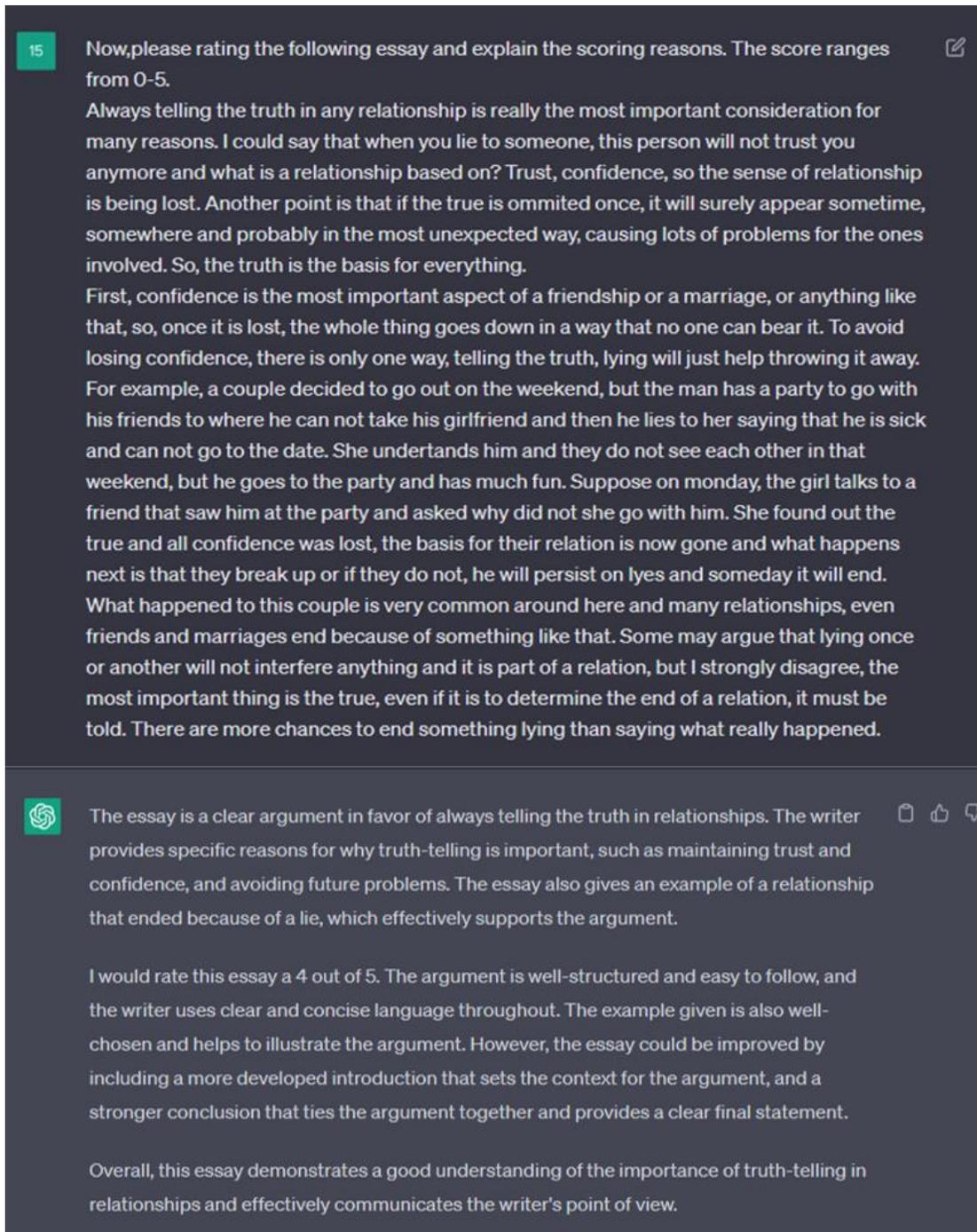

**Fig. 13 Demonstration example.**

4) **Answer-Label Mapping.** In this section, we compare the answers predicted by ChatGPT with the scores outlined in the official guidelines. The highlighted red scores indicate cases where ChatGPT's scores align with the scores from the official standard examples.

**Table 1 presents specific predicted values generated by ChatGPT for the "Basic Task Description + Single Example Score 3" scenario.**

|                        | Score1 | Score2 | Score3              | Score4 | Score5 |
|------------------------|--------|--------|---------------------|--------|--------|
| ChatGPT 1st Prediction | 2      | **2**  | Doesn't make sense  | 3      | 4      |
| ChatGPT 2nd Prediction | 2      | **2**  | Doesn't make sense  | **4**  | 4      |
| ChatGPT 3rd Prediction | 2      | **2**  | Doesn't make sense  | **4**  | 4      |

**(3) Feedback Design**

In the context of feedback design, it is imperative to consider the incorporation of comment feedback. This aspect entails the examination of comment feedback across three distinct scenarios:

Integrated Comment Feedback: The first feedback design integrates comment feedback directly into the prompt information, as demonstrated in the previously discussed prompt design. In this approach, comment feedback is furnished alongside the example, serving as a prompt for ChatGPT's automated essay scoring.

Post-Scoring Comment Feedback: The second feedback design involves the provision of relevant comment feedback to ChatGPT subsequent to its initial scoring prediction based on the provided prompt information. This enables ChatGPT to perform a secondary evaluation of the essay, thereby achieving the desired feedback design effect.

Addressing "Irrelevant Content": The third feedback design focuses on the notion of "irrelevant content." The primary prediction template here comprises the "Basic Task Description" augmented by the "Predicted Score." While natural language processing technology has made substantial strides in detecting typographical errors and grammatical mistakes, the challenge persists in discerning essays that deviate from the intended topic. In particular, essays characterized by a high level of linguistic proficiency but lacking relevance to the assigned topic can significantly mislead the scoring system. The concept of "irrelevant content" centers on ChatGPT's capacity to identify essays with varying themes. To assess this capability, we have selected 32 high-scoring essays, each perfect in its execution, yet divergent in terms of thematic content. These essays serve as a rigorous challenge to ChatGPT's ability to discern thematic variations. The experimental approach involves the deliberate selection of distinct themes and the presentation of high-scoring essays that deliberately stray off-topic. Furthermore, this design accommodates the additional complexity introduced by essays with themes that are either similar or dissimilar but remain off-topic.

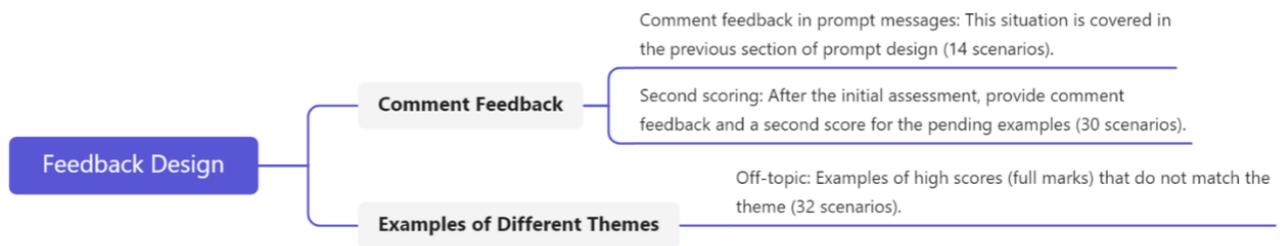

Fig. 14 Specific Components of Feedback Design (76 scenarios)

**(4) Evaluation Criteria**

Lim et al. (2021) conducted a comprehensive review, comparing various commonly used evaluation metrics for assessing the efficiency of Automated Essay Scoring (AES). Their study demonstrated the effectiveness of Quadratic Weighted Kappa (QWK). Therefore, this study adopts QWK as the primary evaluation metric for assessing ChatGPT's automated scoring of TOEFL independent writing tasks.

QWK enhances the treatment of scoring errors through weighted adjustments, which are derived from Kappa statistics. Its definition can be summarized as follows: Initially, a weight matrix W is constructed using Equation 1, where 'i' represents the standard score, 'j' represents the predicted score generated by the ChatGPT model, and 'N' denotes the number of distinct essay score levels. Subsequently, matrices Q and E are constructed to represent the sample counts where the standard score is 'i' and the ChatGPT predicted score is 'j.' Matrix E is the outer product of the standard score and ChatGPT's predicted score vector for TOEFL independent writing tasks. Finally, the QWK value is computed using the constructed matrices E and Q according to Equation 2.

$$W_{i,j} = \frac{(i-j)^2}{(N-1)^2} \quad \text{Equation 1}$$

$$K = 1 - \frac{\sum_{i=1}^{N}\sum_{j=1}^{N} W_{ij} Q_{ij}}{\sum_{i=1}^{N}\sum_{j=1}^{N} W_{ij} E_{ij}} \quad \text{Equation 2}$$

When assessing ChatGPT's results for TOEFL independent writing tasks using QWK, it is important to note that a lower QWK value signifies poorer consistency between standard scores and predicted scores, whereas a higher QWK value indicates stronger consistency. The QWK metric has a range of values from 0 to 1, with specific interpretations detailed in Table 2.

**Table 2 Meaning of QWK Evaluation Metrics**

| QWK Value | ≤ 0.20 | 0.21-0.40 | 0.41-0.60 | 0.61-0.80 | 0.81-1 |
|---|---|---|---|---|---|
| Strength of Agreement | Poor | Fair | Moderate | Strong | Substantial |

**Experimental Data Results**

**(1) Prompt Design Results for TOEFL Independent Writing Tasks by ChatGPT under Different Prompt Information**

Prompt designs were based on the rating elements provided in the official TOEFL guidelines. Through QWK evaluation, the foundational designs of the study might still possess some degree of error, which will be discussed in detail later.

By using the "Basic + Scoring criteria" as the standard data, which outperforms the control group, are highlighted in red.

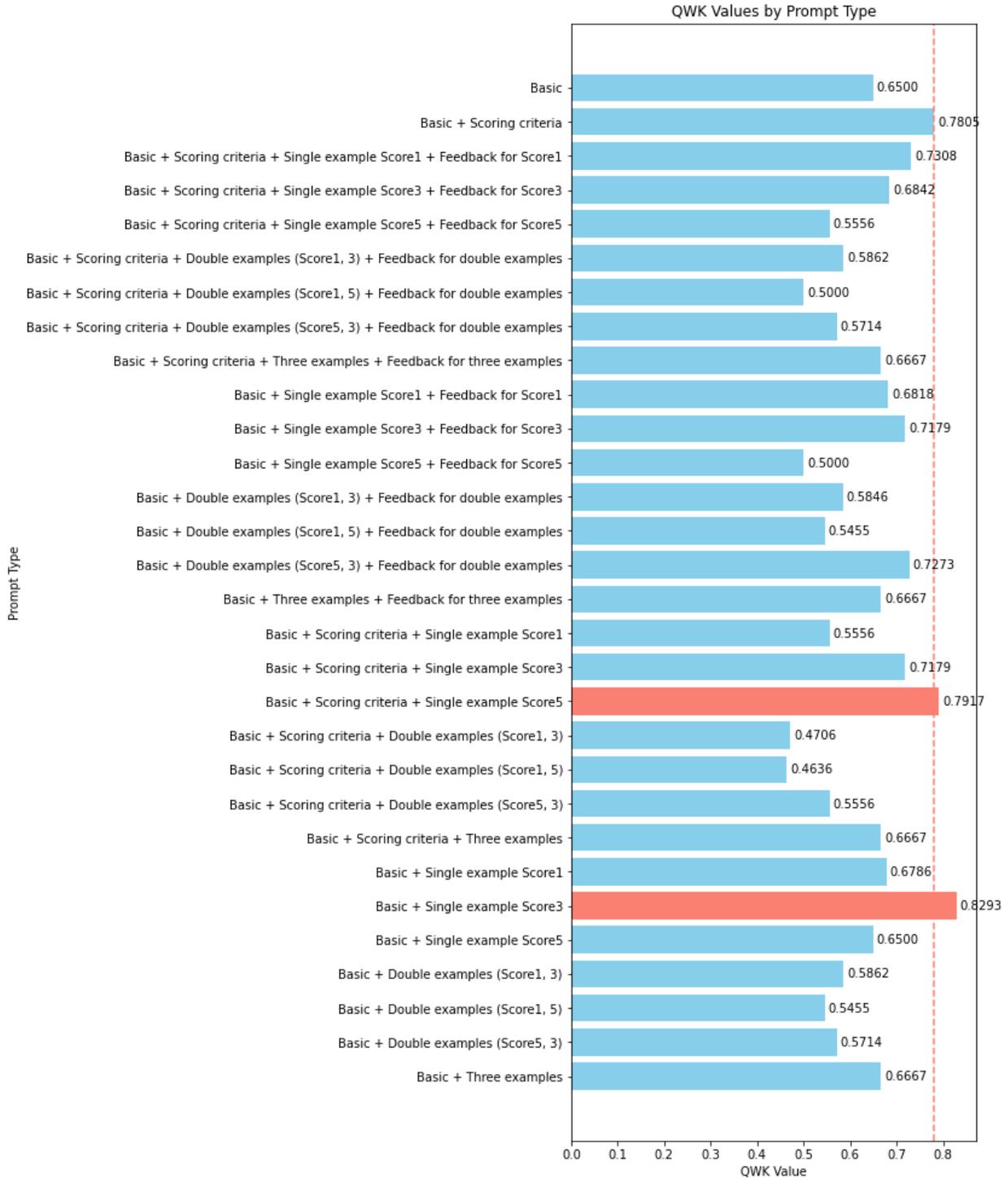

**Fig. 15 Prompt design's bar chart**

As shown in Figure 15: the prompt design data reveals that adding scoring rules to the basic task description can increase the QWK value from the initial baseline of 0.65 to 0.7805.

The prompt information of "basic task description + single example Score3" leads to the best performance of ChatGPT's scoring, with a QWK value of 0.8293. This suggests that scoring rules and examples can enhance ChatGPT's automated scoring and that providing more comprehensive prompt information does not necessarily result in more accurate ChatGPT scoring. The number of examples also affects ChatGPT's scoring, with the performance of a single example being better than that of two or three examples. Furthermore, different prompt design results indicate that zero-shot learning is gradually approaching few-shot learning, which is in line with Zhong et al.'s (2023) conclusion. They suggest that this development is reasonable due to adding prompt design and human feedback to large language models, enabling better understanding of task context and meaning.

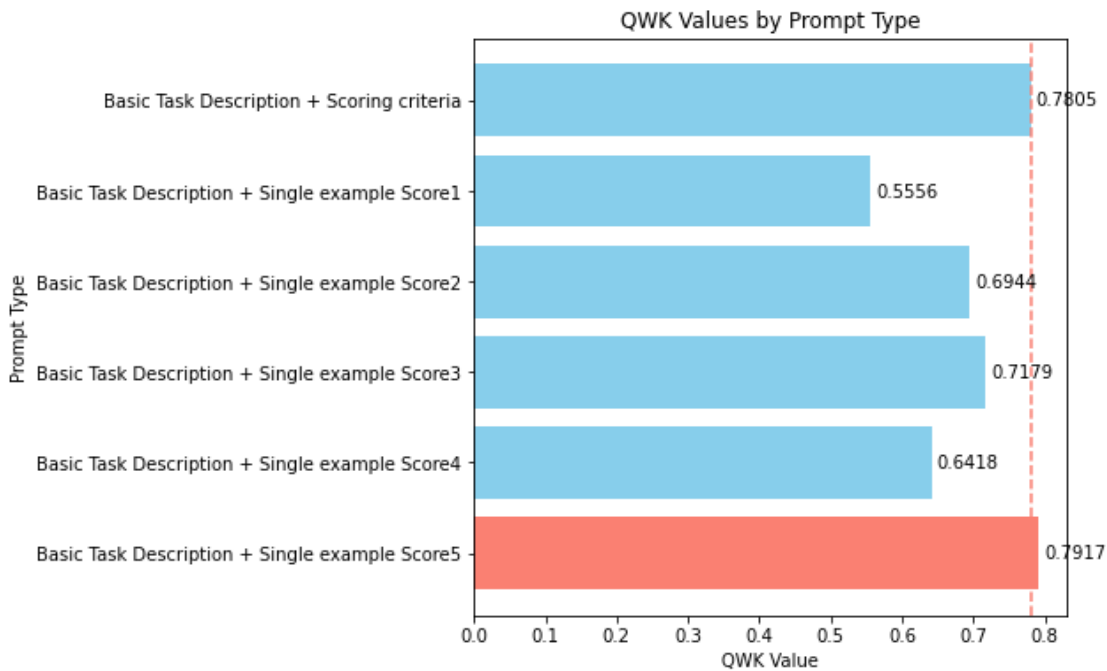

**Fig. 16 Supplementary Experiment1's bar chart**

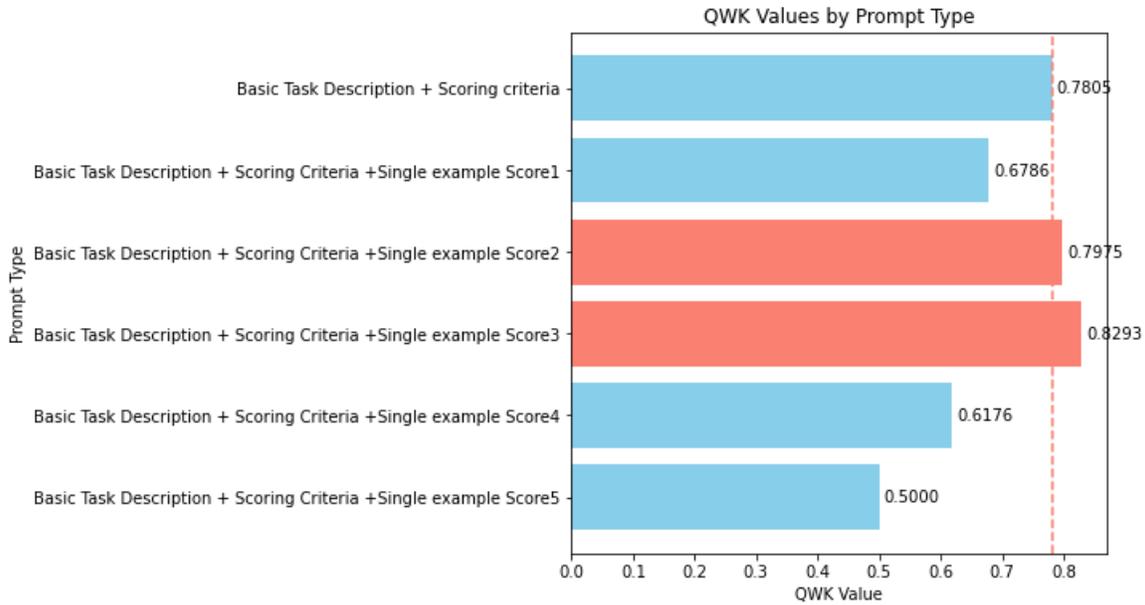

Fig. 17 Supplementary Experiment2's bar chart

In addition, supplementary experiments were conducted for the two highlighted prompt designs in Figure 15 to validate the impact of the selected examples from different score ranges on the scoring results, as shown in Figures 16 and 17. The choice of examples from different score ranges in the prompt information can also lead to varying effects on the article's evaluation.

**(2) Feedback Design Results for TOEFL Independent Writing Tasks by ChatGPT under Different Prompt Information**

In feedback design, there is mostly consistency between ChatGPT's predicted results and the scores given in comment feedback, but some inconsistencies exist. This suggests that ChatGPT possesses contextual learning ability but may exhibit occasional error fluctuations. Experimental data shows that feedback design improves the accuracy of ChatGPT's predicted scores, with resulting QWK values ranging from 0.6 to 1.

Based on the previous feedback design analysis, three distinct hypotheses have emerged. The results of the composition automatic score feedback design experiment conducted with ChatGPT are presented below:

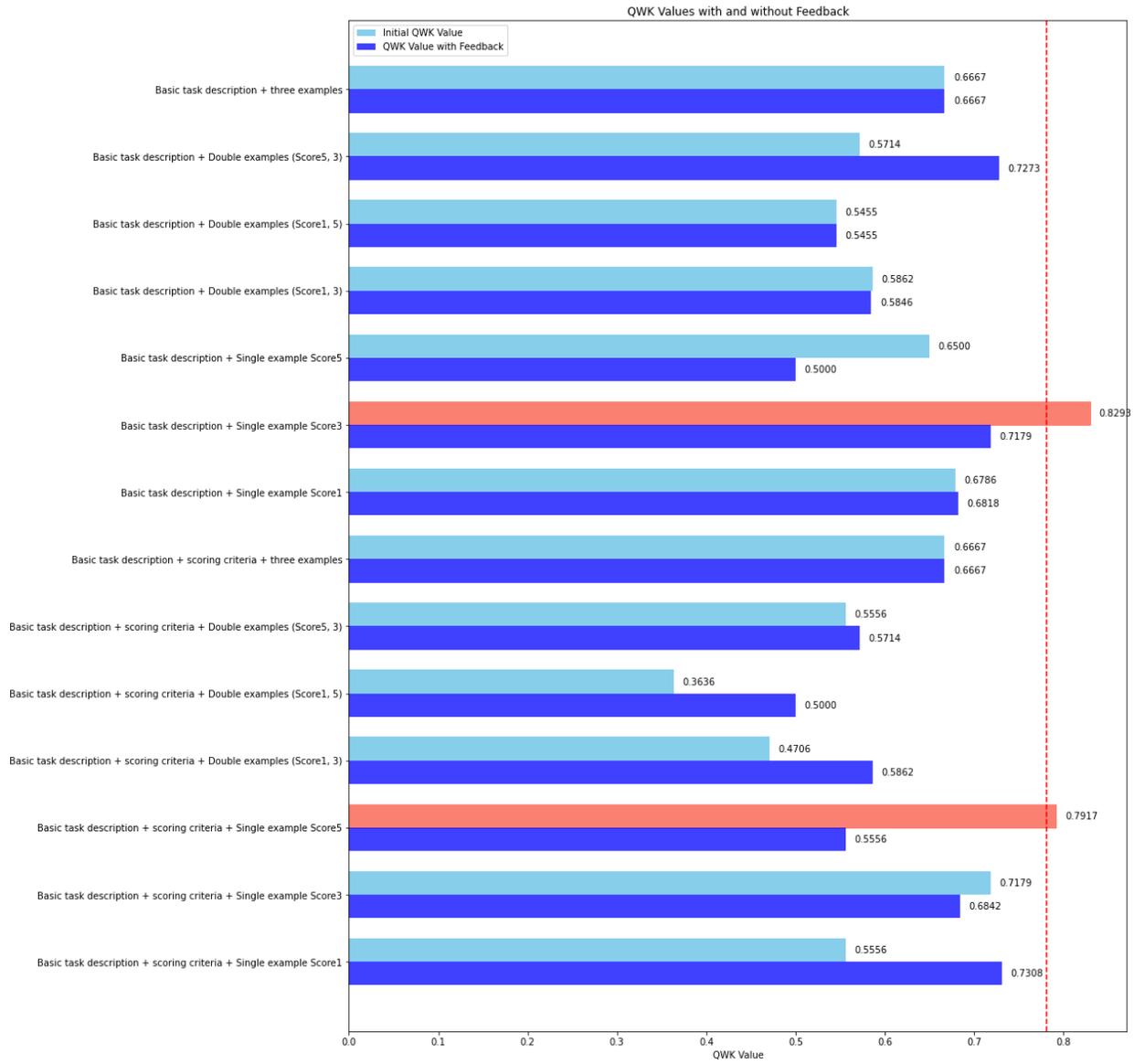

**Fig. 18 Feedback Design1-Integrated Comment Feedback's bar chart**

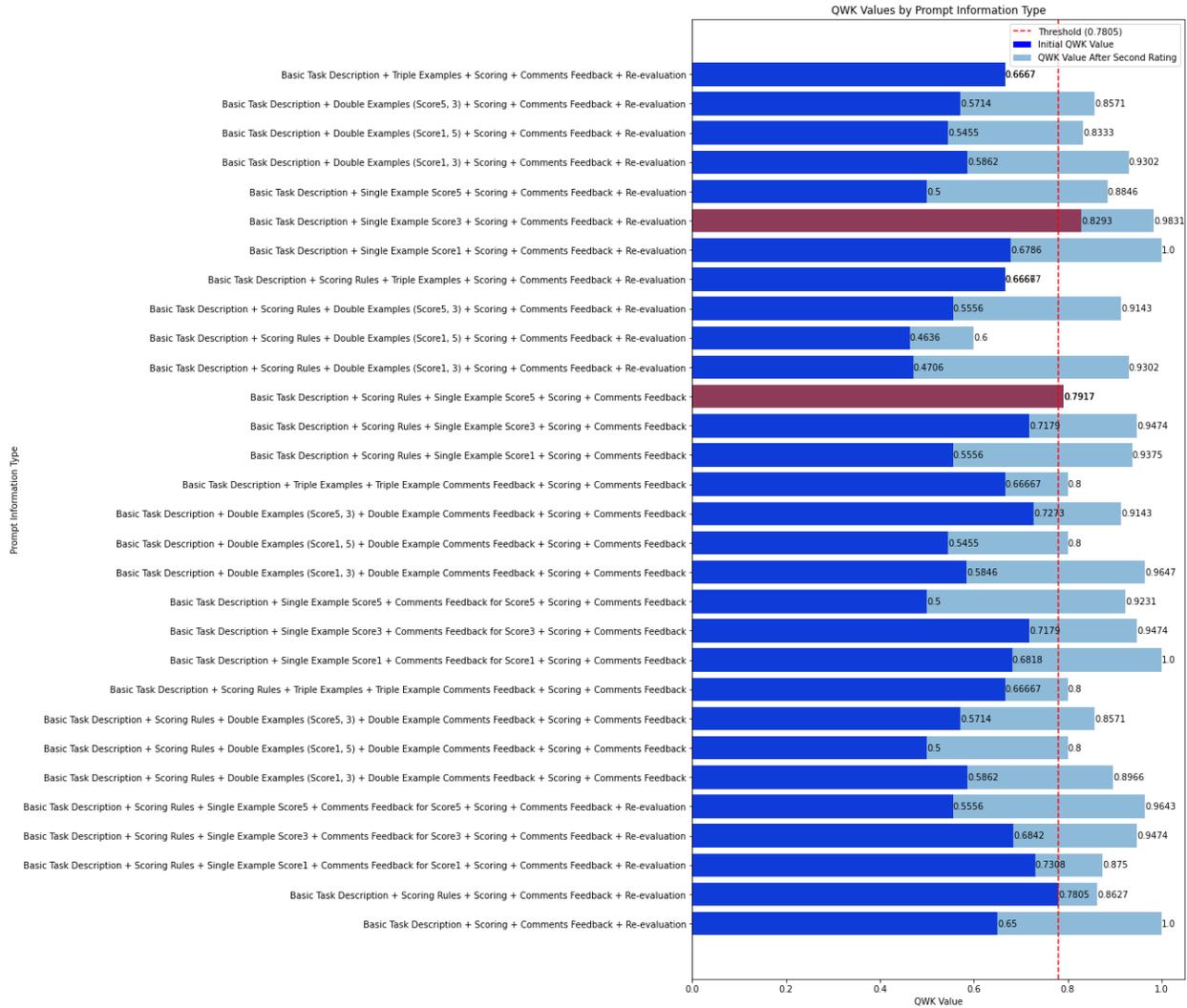

Fig. 19 Feedback Design2-Post-Scoring Comment Feedback's bar chart

Table 3 Feedback Design3 - Addressing "Irrelevant Content" (Prompt Template: Basic Task Description + Predictive Scoring)

| Theme Type | Number of Articles to be Evaluated (unit: articles) | Number of Articles Correctly Identified as "Off-Topic" (unit: articles) |
|---|---|---|
| Life | 4 | 4 |
| Social Friendship | 6 | 6 |
| Teacher Education | 2 | 2 |
| Student Behavior | 2 | 2 |
| School Behavior | 2 | 2 |
| Individual Behavior | 2 | 2 |
| Company Behavior | 2 | 2 |
| Social Change | 2 | 2 |
| Technological Development | 6 | 4 |
| Media | 2 | 0 |
| Work  | 2   | 1      |
| Total | 32  | 27     |
| **Accuracy** | | **84.375%** |

Regarding specific feedback designs, the first type, which uses example-based feedback as prompt information, is suitable for cases where initial predictive results are unsatisfactory. The second type of feedback design, involving targeted secondary scoring of comment feedback, enhances overall predictive results. Lastly, the third type of feedback design considers cases where content is off-topic and finds that ChatGPT can effectively identify differences in the themes of essays, with an accuracy of 84.375%.

**(3) Experimental Data Results for TOEFL Independent Writing Tasks by ChatGPT under Different Theme Types**

In the preceding preliminary experimental section, we have identified two optimal prompt designs for the prompting process: "Basic task description + scoring criteria + 5-point single example + scoring" and "Basic task description + 3-point single example + scoring." To validate the accuracy of our earlier experiments, we consider that the first half of the preliminary experiments has already been conducted on essays from different score ranges, all based on a single topic provided in the TOEFL official guide. The latter half of our experimentation will focus on essays covering a wide range of topics. This entails converting predicted scores of essays from different score ranges on a single topic into scores for high-scoring essays across multiple distinct topics.

To achieve this, our study has collected essays that received full marks or high scores for 11 different TOEFL independent writing tasks, each covering a different topic. We conducted another round of experiments using prompts composed of "Basic task description + 5-point single example + scoring." The experimental results are presented in Table 4.

**Table 4 Evaluation of Predictive Performance by Theme Types (Prompt Template: Basic Task Description + Scoring Criteria + Example of a 5-point Item + Predicted Scores)**

| Theme Type | Number of Articles to be Evaluated (Unit: Articles) | Number of Correct Predictions by ChatGPT (5 points or 4 points) |
|---|---|---|
| Life | 4 | 4 |
| Social Friendship | 6 | 6 |
| Teacher Education | 2 | 2 |
| Student Behavior | 2 | 2 |
| School Behavior | 2 | 2 |
| Individual Behavior | 2 | 1 |
| Company Behavior | 2 | 2 |
| Social Change | 2 | 1 |
| Technological Development | 6 | 4 |
| Media | 2 | 2 |
| Work | 2 | 1 |
| Total | 32 | 27 |
| **Accuracy** | | **84.375%** |

Table 4 presents an evaluation of predictive performance by theme type. It includes the number of articles evaluated for each theme type and the corresponding number of correct predictions by ChatGPT, categorized into 5-point or 4-point accuracy. The total number of articles evaluated sums up to 32, with an overall accuracy rate of 84.375%.

## (4) Overall Results for TOEFL Independent Writing Tasks by ChatGPT

In general, ChatGPT can effectively function as an automated essay scoring tool, capable of discerning differences between writing tasks at various proficiency levels. Experimental calculations yielded Quadratically Weighted Kappa (QWK) values ranging from 0.4636 to 0.82930, demonstrating an 84.375% accuracy in identifying instances where the content deviates from the intended topic. The correct identification rate for articles of different theme types can also reach 84.375%. Those indicated that ChatGPT's essay scoring exhibits a moderate alignment with official standard scoring and is capable of distinguishing to some extent between data related to various topics. Consequently, ChatGPT can be employed for rudimentary automated essay scoring purposes.

## Discussion and Future Outlook

**(1) Discussion**

### 1) Empirical Study Exploring ChatGPT's Application in Automated Essay Scoring.

Thorp (2023) emphasized ChatGPT's primary role in hypothesis generation, experimental design, and result interpretation. Kasneci et al. (2023) discussed the benefits and challenges of using large language models in education from both student and teacher perspectives. Mhlanga (2023) focused on the ethical use of ChatGPT in education. Yogesh et al. (2023) exploring the skills and resources needed for effective use of generative AI. Adiguzel et al. (2023) outlined AI's potential applications in education and associated challenges. Ramesh & Sanampudi (2022) noted variations in ChatGPT's performance across subjects.

This paper takes a microscopic approach, leveraging automated essay scoring to analyze the abilities, limitations, and strengths of large language models in the field of essay scoring through prompt design and feedback design. It's important to note that while this paper presents an empirical verification of the paradigm in the essay scoring domain, the rapid development of large language models and the introduction of new models may impact the specific conclusions drawn here. Therefore, when applying this research paradigm to newer versions of GPT, adjustments should be made accordingly.

When considering the application and promotion of ChatGPT, it is suitable for use in low-stakes examinations, such as classroom essay quizzes. However, it is essential to acknowledge the limitations of ChatGPT's automated scoring in high-stakes scenarios. The experimental results reveal that ChatGPT faces challenges when assigning the lowest and highest scores, displaying a regression effect. To deploy it effectively in high-stakes testing scenarios, further meticulous research is warranted, potentially involving manual scoring to some extent.

**2）Design and Implementation Challenges in ChatGPT's Prompt Information.** In empirical research focused on the application of ChatGPT to automated essay scoring, the evaluation data indicates potential issues in the foundational study design, as outlined below. Initially, it was hypothesized that a more comprehensive incorporation of prompt information during the design phase would result in more precise scoring. However, the results derived from specific prompt design and indicator calculations demonstrate that combinations of distinct prompt information and indicators, such as "basic task description + score" and "basic task description + scoring rules + score," elevate the QWK (Quadratic Weighted Kappa) value from the initial baseline of 0.65 to 0.7805. Subsequent iterations of prompt information combinations did not significantly enhance the efficacy of automated essay scoring. This phenomenon might be attributed to the lingering influence of earlier prompt information in subsequent experimental designs, potentially introducing certain errors. Consequently, fluctuations in QWK values were observed throughout the experiments, and a satisfactory explanation remains elusive within the current scope of research. As a result, the intricate designs applied in ChatGPT's subsequent scoring should be deemed as experimental anomalies, necessitating more precise and comprehensive research to ascertain their accuracy.

Based on the experimental design and findings presented in this paper, it can be inferred that effective and accurate utilization of ChatGPT requires a proficient aptitude for prompt design. While the prompt design process within this paper is relatively exhaustive, ChatGPT's performance does not exhibit remarkable excellence. Several contributing factors can be identified, including ChatGPT's inherent capabilities, potential unidentified shortcomings in prompt design, and imprecise or limited sample data provided within the prompts. Additionally, the implementation phase following prompt design also demands robust engineering skills, as

Empirical Study of Large Language Models as Automated Essay Scoring Tools    34any shortcomings identified during experimentation must be promptly rectified to continually enhance the relevant expertise and practical experience.

3) **Harnessing Domain Expertise for Precision in Fine-Tuning Prompts and Feedback in Few-Shot Learning.** The rapid advancement of natural language processing has led to the widespread utilization of automated scoring systems across diverse domains. This study explores the utilization of domain expert knowledge to enhance few-shot learning through the refinement of prompt and feedback designs, with the ultimate aim of improving ChatGPT's automated essay scoring capabilities. By drawing upon the comprehensive set of rating criteria delineated in the official TOEFL guidelines for independent writing tasks, which inherently encapsulate a wealth of expertise within the English language domain, this investigation has implemented ChatGPT's automated essay scoring in conjunction with carefully crafted few-shot prompts.

By extensively tapping into domain expert knowledge, the meticulous construction of prompts and feedback mechanisms serves to facilitate learning from the collective wisdom and experiences of experts, thereby furnishing ChatGPT with more precise and targeted prompts. Furthermore, each interaction with ChatGPT commences with prompts framed in natural language. In the context of human-ChatGPT interactions, the quality of input prompts plays a pivotal role in shaping the quality of ChatGPT's output. Consequently, in the quest to enhance the quality of ChatGPT's outputs, it becomes imperative for humans to leverage the specialized knowledge of domain experts when meticulously constructing prompt information and experimenting with diverse prompts. This iterative process guides ChatGPT towards the production of desired results that align with human feedback. Consequently, the leveraging of domain experts' knowledge in the development of well-crafted prompts and feedback mechanisms stands as a paramount factor in achieving optimal performance.

**(2) Future Outlook**

In this empirical research article, we delve into an examination of the fundamental capabilities of the prominent large language model, ChatGPT, within the domain of automated scoring for English essays. This exploration is facilitated through the creation of various permutations of prompt information. In the forthcoming period, ChatGPT holds promise in its application for the automated assessment of open-ended questions in STEM fields. This application encompasses the provision of scored writing feedback, generation of concise writing summaries, identification of grammatical errors, evaluation of discourse structure quality, and the discernment of refined linguistic expressions.

It is important to note that the landscape of large language models is undergoing rapid evolution, akin to an arms race. Consequently, the research paradigm presented in this article, along with the specific research inquiries posed, can be extrapolated to encompass the latest iterations of GPT, as well as the expansive array of large language models that are emerging from research and development initiatives in China.

Empirical Study of Large Language Models as Automated Essay Scoring Tools	39

**Figures legends**

*Fig. 1 Illustrates the central process of rule-based natural language processing*
*Fig. 2 Illustrates the fundamental methodology of feature-based statistical machine learning for natural language processing*
*Fig. 3 Illustrates the core mechanics of natural language processing employing deep neural networks*
*Fig. 4 The Basic Pretraining + Fine-tuning Process for Natural Language Processing*
*Fig. 5 The Fundamental Process of Natural Language Processing Using Generative Large Language Models*
*Fig. 6 Illustration of the prompt construction process (Liu et al., 2021)*
*Fig. 7 Relationship Mapping Between Answers and Labels (Liu et al., 2021)*
*Fig. 8 Process of Answer Prediction (Liu et al., 2021)*
*Fig. 9 Illustrates the comprehensive process involved in Sentiment Classification Tasks using the Prompt Learning Method (Liu et al., 2021)*
*Fig. 10 Distribution of the Number of Test Articles for Different Themes*
*Fig. 11 Specific Composition of Prompt Design (30 Scenarios)*
*Fig. 12 Demonstration example.*
*Fig. 13 Demonstration example.*
*Fig. 14 Specific Components of Feedback Design (76 scenarios)*
*Fig. 15 Prompt design's bar chart*
*Fig. 16 Supplementary Experiment1's bar chart*
*Fig. 17 Supplementary Experiment2's bar chart*
*Fig. 18 Feedback Design-Integrated Comment Feedback's bar chart*
*Fig. 19 Feedback Design-Post-Scoring Comment Feedback's bar chart*

**Table**

*Table 1 presents specific predicted values generated by ChatGPT for the "Basic Task Description + Single Example Score 3" scenario.*
*Table 2 Meaning of QWK Evaluation Metrics*
*Table 3 Feedback Design - Addressing "Irrelevant Content" (Prompt Template: Basic Task Description + Predictive Scoring)*
*Table 4 Evaluation of Predictive Performance by Theme Types (Prompt Template: Basic Task Description + Scoring Criteria + Example of a 5-point Item + Predicted Scores)*

## Data availability

All data for this study are summarized in tables and are accessible by using this study's reference list, which provides paths to data available at appendix.

## Competing interests

The authors declare no competing interests.

**Ethical approval**

This article does not contain any studies with human participants performed by any of the authors

**Informed consent**

This article does not contain any studies with human participants performed by any of the authors.